%% file: _main.tex
\input{_constants}
\arxiv

\pdfoutput=1
\documentclass[10pt,twocolumn,letterpaper]{article}
\input{cvpr_header}
\begin{document}
\title{\paperTitle}
\author{\authorBlock}
\maketitle

\input{00_abstract}
\input{01_intro}
\input{02_related}
\input{03_method}
\input{04_experiments}
\input{05_conclusion}
\section{Acknowledgement}
This work was conducted by Center for Applied Research in Artificial Intelligence (CARAI) grant funded by DAPA and ADD (UD230017TD).
{\small
\bibliographystyle{ieee_fullname}
\bibliography{references}
}

\ifarxiv \clearpage \input{06_appendix} \fi

\end{document}

%% file: _constants.tex
\def\cvprPaperID{11528}
\def\confName{ICCV}
\def\confYear{2023}

\def\paperTitle{Audio-Visual Glance Network for Efficient Video Recognition}

\def\authorBlock{
    Muhammad Adi Nugroho \qquad Sangmin Woo \qquad Sumin Lee \qquad Changick Kim \\
    Korea Advanced Institute of Science and Technology (KAIST) \\
    {\tt\small \{madin, smwoo95, suminlee94, changick\}@kaist.ac.kr}
}

\newif\ifreview 
\newif\ifarxiv \newcommand{\arxiv}{\arxivtrue}
\newif\ifcamera 
\newif\ifrebuttal 

%% file: cvpr_header.tex
\ifreview \usepackage[review]{cvpr} \fi
\ifarxiv \usepackage[pagenumbers]{cvpr} \fi
\ifrebuttal \usepackage[rebuttal]{cvpr} \fi
\ifcamera \usepackage{cvpr} \fi

\usepackage{graphicx}
\usepackage{amsmath}
\usepackage{amssymb}
\usepackage{booktabs}

\input{_macros}  
\input{_our_changes}  

\usepackage{xr-hyper}

\makeatletter
\newcommand*{\addFileDependency}[1]{
  \typeout{(#1)}
  \@addtofilelist{#1}
  \IfFileExists{#1}{}{\typeout{No file #1.}}
}

\makeatother

\usepackage[pagebackref,breaklinks,colorlinks,citecolor=citecolor,linkcolor=linkcolor,bookmarks=false]{hyperref}
\usepackage[capitalize]{cleveref}
\crefname{section}{Sec.}{Secs.}
\Crefname{section}{Section}{Sections}
\Crefname{table}{Table}{Tables}
\crefname{table}{Tab.}{Tabs.}

%% file: _macros.tex
\usepackage{times}
\usepackage{microtype}
\usepackage{epsfig}
\usepackage[table,xcdraw]{xcolor}
\usepackage{caption}
\usepackage{float}
\usepackage{placeins}
\usepackage{color, colortbl}
\usepackage{stfloats}
\usepackage{enumitem}
\usepackage{tabularx}
\usepackage{xstring}
\usepackage{multirow}
\usepackage{xspace}
\usepackage{url}
\usepackage{xcolor}
\usepackage[hang,flushmargin]{footmisc}
\usepackage{enumitem}

\ifcamera \usepackage[accsupp]{axessibility} \fi

% ...

\definecolor{skyblue}{RGB}{21,115,176}
\definecolor{orange}{RGB}{230,124,37}

\ifarxiv  \fi

\newcommand{\R}[1]{{%
    \textbf{%
        \ifstrequal{#1}{1}{\textcolor{red}{R#1}}{%
        \ifstrequal{#1}{2}{\textcolor{blue}{R#1}}{%
        \ifstrequal{#1}{3}{\textcolor{magenta}{R#1}}{%
        \ifstrequal{#1}{4}{\textcolor{teal}{R#1}}{%
                           \textcolor{cyan}{R#1}%
        }}}}%
    }%
}}

%% file: _our_changes.tex
\usepackage{times}
\usepackage{epsfig}
\usepackage{graphicx}
\usepackage{amsmath}
\usepackage{amssymb}
\usepackage{booktabs}
\usepackage{comment}
\usepackage{color}
\usepackage{xcolor}
\usepackage[numbers,sort]{natbib}
\usepackage{multirow}
\usepackage{algorithm}
\usepackage{algorithmic}
\usepackage{dsfont}
\usepackage{bm}  
\usepackage{xspace}
\usepackage{colortbl}
\usepackage{diagbox}
\usepackage{enumitem}
\usepackage{placeins}
\usepackage{footnote}
\usepackage[title]{appendix}
\usepackage{tikz}
\usepackage{amsfonts}       
\usepackage{bbding}
\usepackage{nicefrac}       
\usepackage{wrapfig}
\usepackage{tabulary}
\usepackage{relsize}
\newlength\abovesecmargin
\newlength\belowsecmargin
\newlength\abovesubsecmargin
\newlength\belowsubsecmargin
\newlength\abovesubsubsecmargin
\newlength\belowsubsubsecmargin
\newlength\paramargin
\newlength\abovetabcapmargin
\newlength\belowtabcapmargin
\newlength\abovefigcapmargin
\newlength\belowfigcapmargin

\makeatletter\renewcommand\paragraph{\@startsection{paragraph}{4}{\z@}
  {.5em \@plus1ex \@minus.2ex}{-.5em}{\normalfont\normalsize\bfseries}}\makeatother

\makeatletter
\DeclareRobustCommand\onedot{\futurelet\@let@token\@onedot}
\def\@onedot{\ifx\@let@token.\else.\null\fi\xspace}

\def\eg{\emph{e.g}\onedot} 
\def\ie{\emph{i.e}\onedot}

\makeatother

\newlength\savewidth

\newcolumntype{x}[1]{>{\centering\arraybackslash}p{#1pt}}
\newcolumntype{y}[1]{>{\raggedright\arraybackslash}p{#1pt}}
\newcolumntype{z}[1]{>{\raggedleft\arraybackslash}p{#1pt}}

\definecolor{Gray}{gray}{0.9}
\definecolor{Green}{rgb}{0.2, 0.7, 0.1}
\definecolor{Orange}{rgb}{0.8, 0.5, 0.2}
\definecolor{Yellow}{RGB}{255, 192, 0}
\definecolor{todo}{rgb}{0.8, 0.4, 0.2}
\definecolor{citecolor}{HTML}{0071BC}
\definecolor{linkcolor}{HTML}{ED1C24}
\definecolor{plus}{HTML}{0071bc}
\definecolor{minus}{RGB}{153,10,10}
\definecolor{quickdraw}{rgb}{.5, .0, .5}
\definecolor{tuberlin}{rgb}{0, 0.3, 0.8}
\definecolor{sketchy}{rgb}{0, .5, 0}

\usepackage{pifont}
\newcommand{\cmark}{\ding{51}}
\newcommand{\xmark}{\ding{55}}

\newcommand{\up}{\bf \fontsize{10}{42} \color{plus}{$\uparrow$}}
\newcommand{\down}{\bf \fontsize{8}{42}\selectfont \color{minus}{$\downarrow$}}

\renewcommand{\Sigma}{\mathfrak{S}}


%% file: 00_abstract.tex
\begin{abstract}
Deep learning has made significant strides in video understanding tasks, but the computation required to classify lengthy and massive videos using clip-level video classifiers remains impractical and prohibitively expensive.
To address this issue, we propose Audio-Visual Glance Network (AVGN), which leverages the commonly available audio and visual modalities to efficiently process the spatio-temporally important parts of a video. AVGN firstly divides the video into snippets of image-audio clip pair and employs lightweight unimodal encoders to extract global visual features and audio features. To identify the important temporal segments, we use an Audio-Visual Temporal Saliency Transformer (AV-TeST) that estimates the saliency scores of each frame. To further increase efficiency in the spatial dimension, AVGN processes only the important patches instead of the whole images. We use an Audio-Enhanced Spatial Patch Attention (AESPA) module to produce a set of enhanced coarse visual features, which are fed to a policy network that produces the coordinates of the important patches. This approach enables us to focus only on the most important spatio-temporally parts of the video, leading to more efficient video recognition. Moreover, we incorporate various training techniques and multi-modal feature fusion to enhance the robustness and effectiveness of our AVGN. By combining these strategies, our AVGN sets new state-of-the-art performance in multiple video recognition benchmarks while achieving faster processing speed.
\end{abstract}

%% file: 01_intro.tex
\section{Introduction}
\begin{figure}[t]
\centering
    \includegraphics[width=\columnwidth]{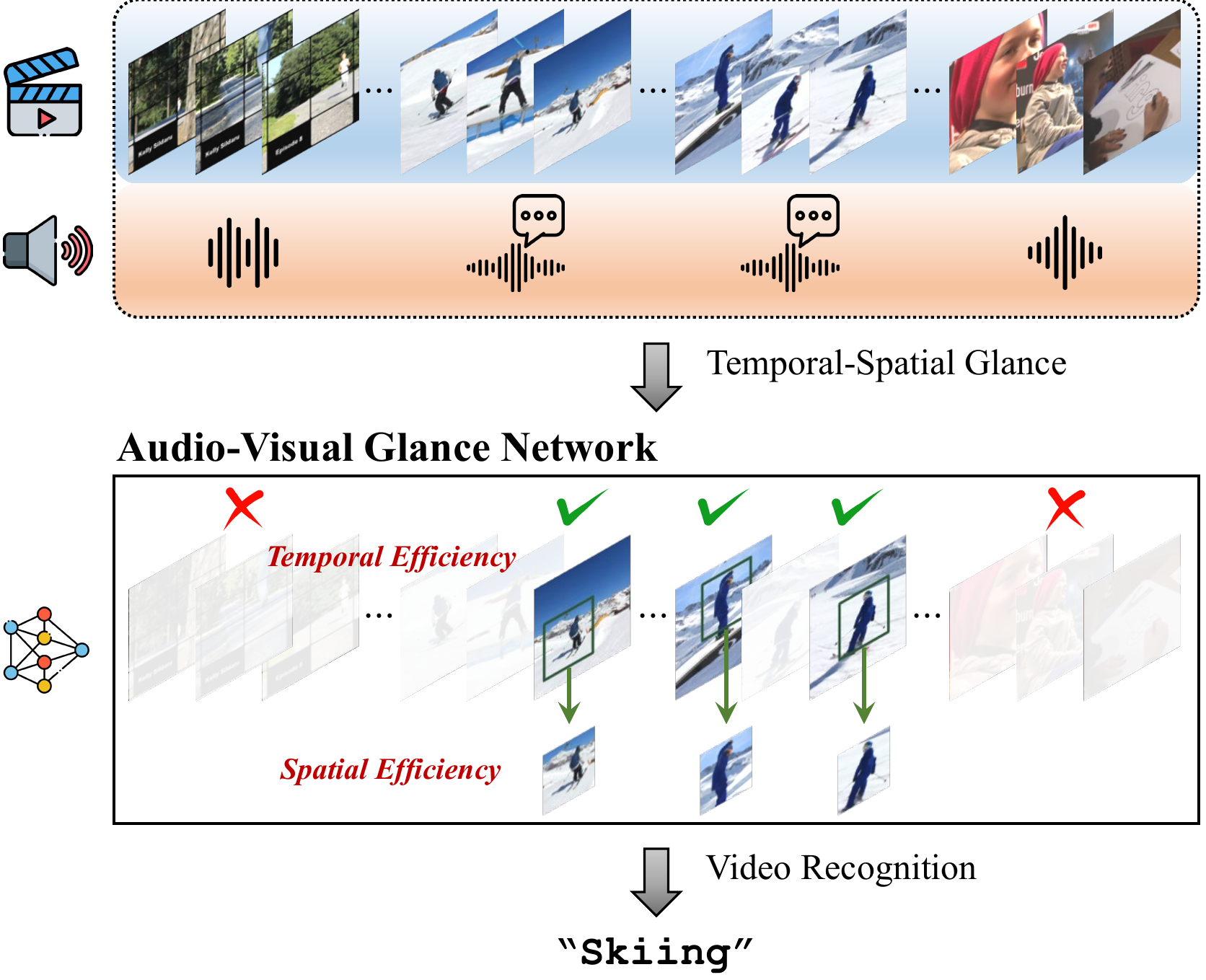}
    \vspace{-0.25in}
    \caption{\textbf{Audio-Visual Glance Network (AVGN)} performs efficient video recognition by processing only a few highly salient frames based on the audio and visual information. Then, it determines and extracts the important spatial area in those frames to construct even more compact video representations.}
    \label{fig:overview}
\vspace{-0.25in}
\end{figure}
The exponential growth of diverse video contents, particularly those involving human-related actions, has prompted the development of deep learning algorithms~\cite{Carreira2017_I3D,Tran2018_C3D,Xiao2020_AVSlowFast,Feichtenhofer2019-SlowFast} that can effectively process and understand such data.
Video recognition can bring benefits to multiple fields, such as sports for performance analysis~\cite{Chen2021_SportRecog}, military for situational awareness~\cite{Geraldes2019_UAVDefense}, transportation for traffic monitoring~\cite{bortnikov2020_accident}, security for threat detection~\cite{sultani2018real} and surveillance for public safety~\cite{dave2022_gabriellav2Surveillance}.
However, the current state-of-the-art methods often require high computational costs, especially when dealing with lengthy and heavy videos, hindering their practical application in real-world scenarios.
This has led to an increasing demand for efficient video understanding methods~\cite{Sun2022_HARReview}.

To address this issue, various approaches have been proposed, such as developing efficient and lightweight architecture ~\cite{Kondratyuk2021_MoViNets}, or adaptively selecting only the most informative subset of given videos~\cite{Wu2022_AdaFrame,Ghodrati2021_frameexit,Zheng2020_DynamicSamplingNetworks}, or training a policy network using policy gradient~\cite{Sun2021_VideoIQ,Meng2020_AR-net}. On the other hand, some approaches manipulate the spatial resolution of input video frames to achieve efficiency, such as extracting only important spatial patches~\cite{Wang2021_adafocus,Wang2022_adafocusv2} or adaptively changing the resolution frames based on the importance~\cite{Meng2020_AR-net}.

From all those approaches, we found that for efficient action recognition, we need to selectively locate and process only the important temporal and spatial location of the video. Intuitively, we need to know \textit{when} and \textit{where} to look at. For example in Fig.~\ref{fig:overview}, to determine the action class ``\texttt{skiing}'', we only need the frames that contain a person riding the ski, not the frames that only show the snowfield or the person doing no action. This inspires us to create a network that can do efficient action recognition by \textit{glancing} through the video to selectively find the important frames in a low-cost manner. Further, we aim to utilize the additional audio modality which is naturally available together with the visual modality in video recording, mimicking the human intuition of skimming through long sequences using both visual and audio cues to find important keyframes. Compared to the visual modality, the action-discriminative features of audio modality are easier to compute~\cite{Sun2022_HARReview}, and it also helps to distinguish actions that are visually similar~\cite{Kazakos2019_EpicFusion}.

We propose Audio-Visual Glance Network (AVGN), a comprehensive framework designed to enhance efficiency in both spatial and temporal dimensions. For temporal efficiency, AVGN aims to make the correct recognition of a video sequence with only a few important frames that actually contain distinctive cues. To achieve this, we construct an Audio-Visual Temporal Saliency Transformer (AV-TeST) that estimates the temporal saliency of a frame using coarse features generated by lightweight audio and visual backbones. Furthermore, to ensure spatial efficiency, we construct an Audio-Enhanced Spatial Patch Attention (AESPA) module that learns the relationship between audio feature sequences and visual features. This module generates audio-enhanced visual features that can be used by a patch extraction network to extract important spatial patches. For each frame, the patch contains only the important area of the image frame and has a lower pixel size compared to the original image. In addition to these modules, we also devise an appropriate feature fusion for classifier input and the training techniques to optimize these modules.

\begin{figure}[t]
\centering
    \includegraphics[width=\columnwidth]{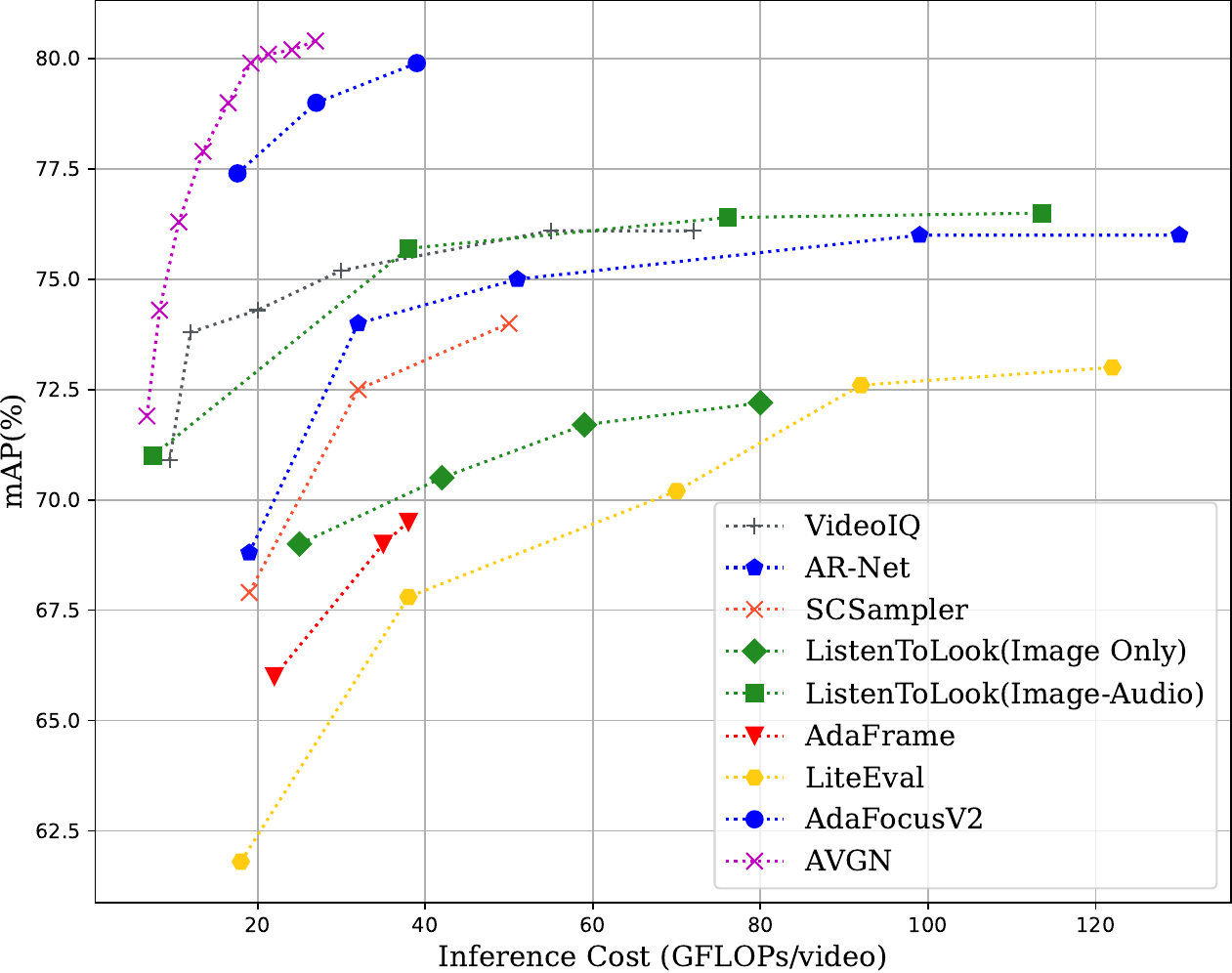}
    \vspace{-0.25in}
    \caption{\textbf{Comparison of performance (mAP(\%)) vs. cost (GFLOPs)} between AVGN and benchmark methods on the ActivityNet dataset~\cite{caba2015_activitynet}. AVGN achieves a pareto optimal by scoring the highest mAP, while also having lower GFLOPs.}
    \label{fig:sota_map}
    \vspace{-0.2in}
\end{figure}

Our experimental results demonstrate that AVGN effectively incorporates audio and visual modalities for efficient action recognition. Our comparison with other state-of-the-art methods in Fig.~\ref{fig:sota_map} shows that AVGN achieves a higher mAP and lower FLOP cost on the ActivityNet dataset, indicating that it achieves pareto optimality. To conclude, our contributions are as follows:
\begin{itemize}[nosep]
    \item We show that incorporating audio modality into the video recognition process can lead to a pareto optimal solution, \ie, improved accuracy without sacrificing efficiency.
    \item Our approach combines audio and visual information to improve temporal and spatial efficiency in a unified manner, and incorporates tailored training strategies that further optimize the performance of our AVGN.
    \item AVGN achieves state-of-the-art performance on multiple video recognition benchmarks as a result of the model building blocks and training techniques.
\end{itemize}

%% file: 02_related.tex
\section{Related Work}
\paragraph{Video recognition.}
Video recognition backbones are commonly used as part of the solution to the action recognition task. Recent advances in deep-learning have led to the development of various techniques, such as C3D~\cite{Tran2018_C3D}, I3D~\cite{Carreira2017_I3D}, and SlowFast~\cite{Feichtenhofer2019-SlowFast}, that can perform this task directly on trimmed video clips or snippets of untrimmed videos~\cite{Wang2017_UntrimmedNets,Wang2019_TSN}. These approaches have been extended to other related areas, such as action localization~\cite{Shou2018_AutoLoc,Wang2017_UntrimmedNets}, multi-modal action recognition~\cite{Woo2023_TowardMisModAR,lee_2023modality,Sun2022_HARReview}, and efficient action recognition~\cite{Liu2020_TEINet,Kim2021_ActRecDynamic,Wu2021_MFVNet,Wu2021_DSANet}. In our work, we develop an efficient video recognition model that can achieve high accuracy on both trimmed and untrimmed videos while minimizing its computational cost.

\paragraph{Efficient video recognition.}
Temporal redundancy due to some frames being visually similar or containing irrelevant backgrounds leads to inefficient video recognition. To address this issue, various methods have been proposed ~\cite{Meng2020_AR-net,Sun2021_VideoIQ,Ghodrati2021_frameexit,Wu2019_MARLSm,Wu2022_AdaFrame,Yeung2016_ActGlimpse,Wu2019_LiteEval, Gowda2021_SMART}. For example, temporal shift module~\cite{Lin2019_TSMVideoUnderstanding} shifts feature maps along the temporal dimension to enable computationally-free temporal connections on top of 2D convolutions. AR-Net~\cite{Meng2020_AR-net} consists of a policy network that decides which resolution to process a frame with, and multiple backbones with various resolutions. VideoIQ~\cite{Sun2021_VideoIQ} contains lightweight policy network that can adjust the quantization precision of frames so that simpler frames are processed with lower precision, while FrameExit~\cite{Ghodrati2021_frameexit} uses an effective deterministic policy network and gating module to find the earliest exiting temporal point in a video sequence. MGSampler~\cite{Zhi2021_MGSampler} and AdaFrame~\cite{Wu2022_AdaFrame} use LSTMs to adaptively decide which frame to process next, while AdaFuse~\cite{Meng2021_AdaFuse} dynamically chooses whether to reuse extracted features or fuse current and past features. The DSN~\cite{Zheng2020_DynamicSamplingNetworks} framework contains a sampling module and a policy maker to perform clip selection, forwarding only important frames to the classification module. In parallel directions, there are models that focus on reducing spatial redundancy. For example, AdaFocus~\cite{Wang2021_adafocus,Wang2022_adafocusv2} uses a lightweight feature extractor and policy network to extract only the most important area or patch from an image, which is then fed into a heavier visual network with lower image resolution. AdaFocusV1~\cite{Wang2021_adafocus} uses reinforcement learning to train the policy network, while AdaFocusV2~\cite{Wang2022_adafocusv2} improves on this by utilizing a bilinear interpolation method to enable backpropagation along with their own training procedure. Our proposed network aims to reduce both temporal and spatial redundancy. For temporal redundancy reduction, we employ a multimodal transformer to assess and select only the most salient frames. Meanwhile, to address spatial redundancy we crop spatially only the most important areas for processing.

\begin{figure*}[t]
\centering
    \includegraphics[width=\textwidth]{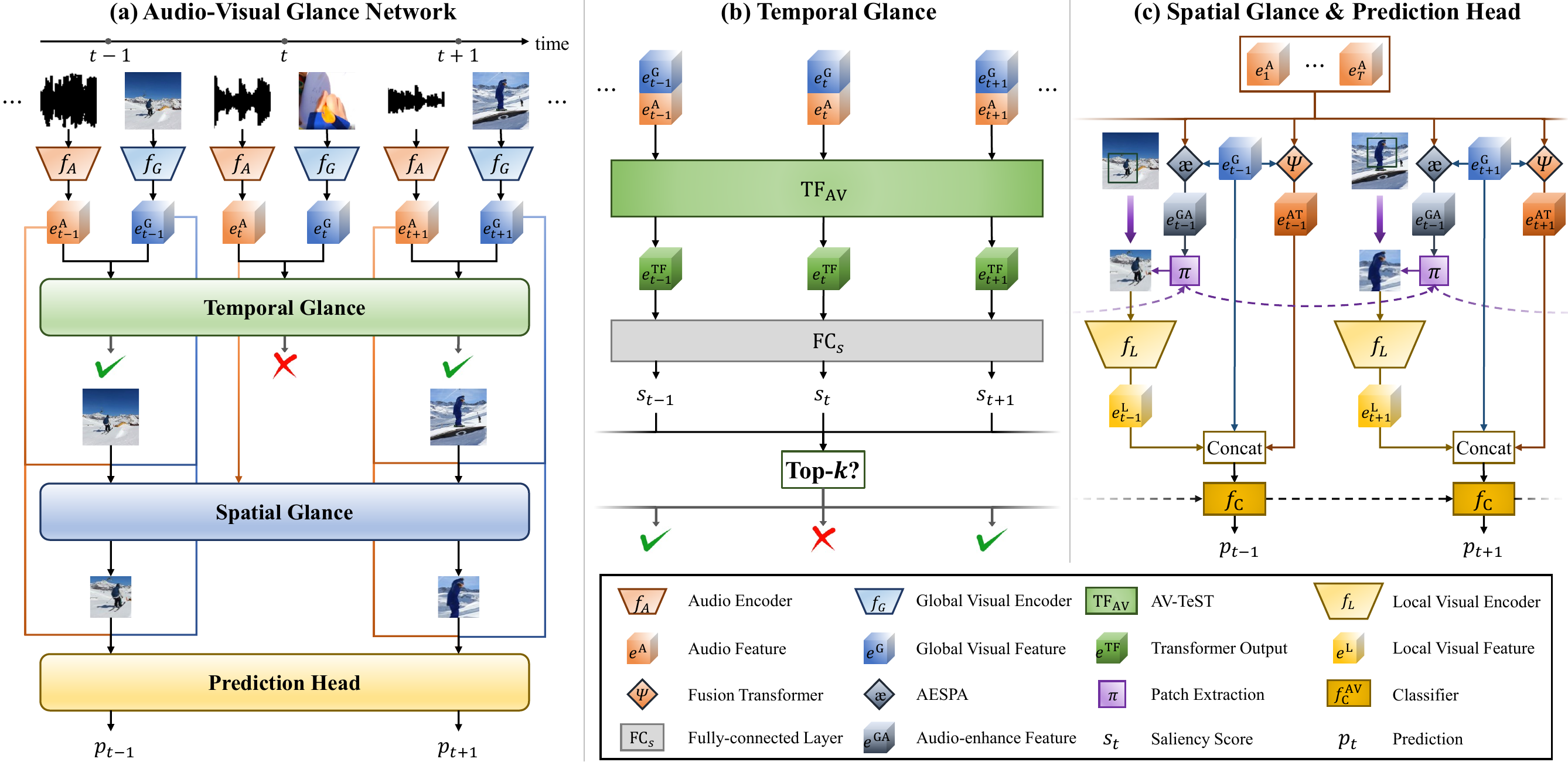}
    \vspace{-0.25in}
    \caption{
    \textbf{Overview of AVGN.}
    \textbf{(a)} At each time step $t$, we extract the features $\emph{e}_t^{\mbox{\scriptsize{A}}}$ and $\emph{e}_t^{\mbox{\scriptsize{G}}}$ using the encoders $f_A$ and $f_G$ from the video. These features are then used for the temporal and spatial glance.
    \textbf{(b)} In temporal glance stage, Audio-Visual Temporal Saliency Transformer (AV-TeST) $\mbox{TF}_{\mbox{\scriptsize AV}}$ generates features $\emph{e}_t^{\rm TF}$, which are then processed with ${\rm FC}_s$ to produce a temporal saliency score $s_t$. Only the top-$k$ salient frames are passed through the spatial glance stage.
    \textbf{(c)} In spatial glance stage, Audio-Enhanced Spatial Patch Attention (AESPA) module $æ$ (see Fig.~\ref{fig:fuse} for more details) enhances the global visual features $\emph{e}_t^{\mbox{\scriptsize{G}}}$ with the sequence of audio features $\{\emph{e}_1^{\mbox{\scriptsize{A}}}, \cdots , \emph{e}_T^{\mbox{\scriptsize{A}}}\}$, resulting in audio-enhanced visual features $\emph{e}_t^{\mbox{\scriptsize{GA}}}$ for the patch extraction network $\pi$ to crop the most important areas of each frame.
    The patches are then fed to the local visual encoder $f_{\mbox{\scriptsize L}}$, which is heavier than $f_G$, to extract features $\emph{e}_t^{\mbox{\scriptsize{L}}}$. Additionally, we use an audio fusion transformer $\psi$ that selectively attends to the relevant parts of the audio feature sequence based on the global visual feature of the current time step $\emph{e}_t^{\mbox{\scriptsize{G}}}$, producing $\tilde{e}_t^{\mbox{\scriptsize{A}}}$. Finally, we concatenate all the extracted features and feed them to the video classifier module $f_c$ to output the prediction $p_t$.
    }
    \vspace{-0.2in}
    \label{fig:AVGN_scheme}
\end{figure*}

\paragraph{Audio in video understanding.}
In video understanding tasks, it has been demonstrated that incorporating additional modalities can improve performance beyond merely using visual modality alone~\cite{Plizzari2022_DomGenAV,Lee2021_AVFuseWSTAL}.
Audio is easy to acquire, requires low computational cost, and has distinctive characteristics.
State-of-the-art video recognition models such as MM-ViT~\cite{Chen2022_MMViT} have incorporated multiple modalities, utilizing cross-modal self-attention blocks, which can handle a large number of multimodal spatio-temporal tokens. Another approach, ListenToLook~\cite{Gao2020_ListenToLook} efficiently integrates the audio modality into an action recognition module by selectively sampling important frames and building a lightweight image-audio pair to mimic a more expensive clip-based model. SCSampler~\cite{Korbar2019_SCSampler} combines both video and audio modalities by utilizing their saliency scores, which are obtained after processing with a lightweight clip sampler network. Our work also leverages the power of audio as an efficient modality for identifying important frames and areas in a video.

%% file: 03_method.tex
\begin{figure}[t]
\centering
    \includegraphics[width=\columnwidth]{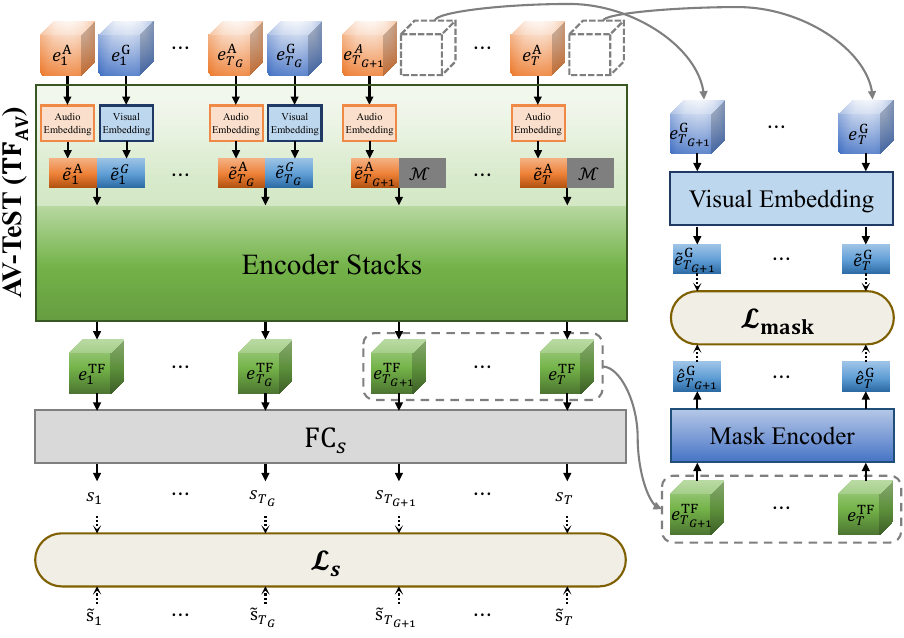}
    \vspace{-0.25in}
    \caption{
    \textbf{AV-TeST components and auxiliary losses.}The saliency loss $\mathcal{L}_s$ is used to train the model to predict the saliency score $s_t$ that matches the confidence score $\tilde{s}_{t}$ of each frame. The masked reconstruction loss $\mathcal{L}_{\mathrm{mask}}$ is used to enhance the robustness of the AV-TeST by partially masking the visual features $\hat{e}^{\mbox\scriptsize{G}}_t$ from frame index $T_{G+1}$ up to the last frame $T$.
    We calculate the L2 loss between the embedded visual features $\tilde{e}^{\mbox\scriptsize{G}}_t$ and the reconstructed visual features $\hat{e}^{\mbox\scriptsize{G}}_t$ extracted by a mask encoder using the output features $e^{\rm TF}_t$ of AV-TeST.
    }
    \vspace{-0.2in}
    \label{fig:AVTST_train}
\end{figure}

\section{Audio-Visual Glance Network}
The overview of AVGN is shown in Fig.~\ref{fig:AVGN_scheme}. Given a video dataset $\mathcal{D}=\{(\boldsymbol{V}_n,\boldsymbol{A}_n,y_n)\}^N_{n=1}$ where each sample contains a sequence of video frames $\boldsymbol{V}=\{v_1,..., v_t,...v_T\}$ $\subset \mathbb{R}^{3 \times H_0 \times W_0}$ emporally paired with a sequence of audio spectrograms $\boldsymbol{A}=\{a_1,...,a_t,...,a_T\}$ $\subset \mathbb{R}^{1 \times H_A \times W_A}$. Our goal is to correctly classify videos into their associated labels $y_n = \{0,1\}^{C}\in \mathbb{R}^{C}$. Our model consists of an audio encoder $f_A$, a coarse global visual encoder $f_G$, a finer local level visual encoder $f_L$, an Audio-Visual Temporal Saliency Transformer (AV-TeST) module $\mbox{TF}_{\mbox{\scriptsize AV}}$, Audio-Enhanced Spatial Patch Attention (AESPA) module $æ$, audio fusion transformer $\psi$, and a spatial patch extraction network $\pi$. 

\subsection{Temporal Glance}
The first stage of our network is to glance over the video sequence to find important temporal locations. The process starts with extracting audio and visual features using the lightweight encoders, $f_A$ and $f_G$, respectively so that,
\begin{equation}
  \emph{e}_t^{\mbox{\scriptsize{A}}} = f_A(a_t), \ \emph{e}_t^{\mbox{\scriptsize{G}}} = f_G(v_t),
  \label{eq:feature_extraction}
\end{equation}
where $\emph{e}_t^{\textrm{G}}\in\mathbb{R}^{D_{\textrm{G}}\times H_{\textrm{G}}\times W_{\textrm{G}}},\emph{e}_t^{\textrm{A}}\in\mathbb{R}^{D_{\textrm{U}}\times H_{\textrm{U}}\times W_{\textrm{U}}}$.
These encoders are designed to have low computational costs to prevent overburdening the glance process. We execute temporal glance using the AV-TeST which consists of modality embedding layers, and a stack of transformer encoders~\cite{Vaswani2017_attuneed} (see Fig.~\ref{fig:AVTST_train}). AV-TeST learns the temporal relationship among the audio-visual pairs, transforming them into audio-visual features ($\textbf{e}_{1:T}^{\mbox{\scriptsize{TF}}}$). The process starts with averaging audio and visual features across their spatial dimension and then we use the embedding layers to produce audio tokens $\tilde{\emph{e}}_t^{\textrm{A}}\in\mathbb{R}^{D_{\textrm{AV}}}$ and visual tokens $\tilde{\emph{e}}_t^{\textrm{G}}\in\mathbb{R}^{D_{\textrm{AV}}}$. We then concatenate audio and global visual tokens from each time step and feed the resulting sequence into the Audio-Visual Temporal Saliency Transformer (AV-TeST) encoder module $\mbox{TF}_{\mbox{\scriptsize AV}}$.
\begin{equation}
\textbf{e}_{1:T}^{\mbox{\scriptsize{TF}}} = \mbox{TF}_{\mbox{\scriptsize AV}}\left(\{[\tilde{\emph{e}}_t^{\textrm{G}},\tilde{\emph{e}}_t^{\textrm{A}}], \cdots,[\tilde{\emph{e}}_t^{\textrm{G}},\tilde{\emph{e}}_t^{\textrm{A}}],\cdots,[\tilde{\emph{e}}_t^{\textrm{G}},\tilde{\emph{e}}_t^{\textrm{A}}]\}\right),
\end{equation}
where $[\cdot,\cdot]$ denotes concatenation operation. Next, we pass each audio-visual feature at time $t$ through a fully-connected layer $\mbox{FC}_{s}$ to obtain a saliency score.
\begin{equation}
s_t = \mbox{FC}_s(\emph{e}^{\mbox{\scriptsize{TF}}}_t).
\end{equation}
Frames with high saliency scores are considered relevant for video recognition, while frames with low saliency scores are considered non-relevant. We process only the $k$ frames with the highest saliency scores for efficient inference.

\subsection{Spatial Glance}
\begin{figure}[t]
\centering
    \includegraphics[width=\columnwidth]{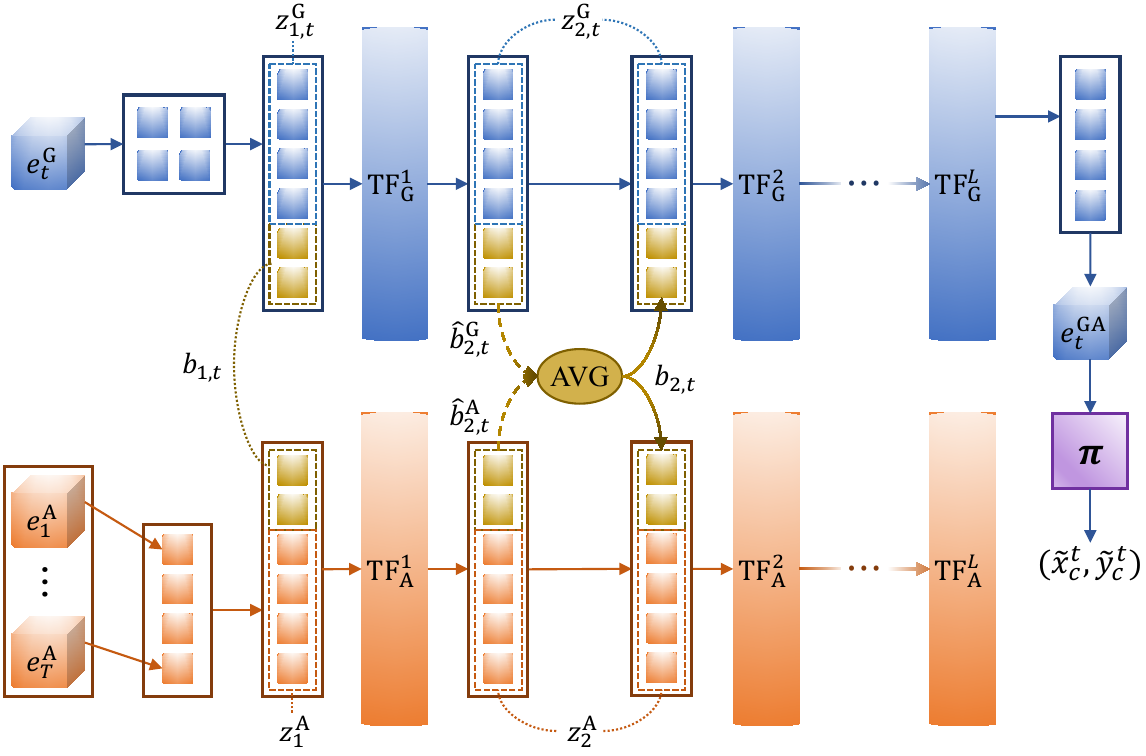}
    \vspace{-0.3in}
    \caption{
    \textbf{The AESPA module} enhances visual features with a sequence of highly correlated audio features through a stack of transformers~\cite{Vaswani2017_attuneed}.
    First, the global visual feature $e^{\rm G}_t$ is flattened, and a sequence of audio features $\{e^{\rm A}_1,\cdots,e^{\rm A}_T\}$ is pooled, flattened, and stacked along the temporal axis.  Bottleneck tokens $b_{1,t}$ are then appended. Next, these augmented features are passed through a series of transformer encoders $\rm{TF}$. After each encoder, the bottleneck tokens are averaged and appended to the audio and visual tokens again. The output visual token of the last transformer layer is reshaped back to match its original shape, resulting in $e^{\mbox{\scriptsize{GA}}}_t$.
    Finally, the patch extraction network $\pi$ utilizes $e^{\mbox{\scriptsize{GA}}}_t$ to produce the visual patch center coordinates $(\tilde{x}^t_c,\tilde{y}^t_c)$.}
    \label{fig:fuse}
    \vspace{-0.2in}
\end{figure}

In the second stage of AVGN, we aim to identify important spatial patches in the video frames. We employ a recurrent patch extraction network $\pi$ inspired by~\cite{Wang2022_adafocusv2}, which utilizes a set of audio-enhanced visual tokens $e^{\mbox{\scriptsize GA}}_{t}$ produced by our proposed Audio-Enhanced Spatial Patch Attention (AESPA) module (see Fig.~\ref{fig:fuse}). AESPA enhances the visual feature of each frame by the whole sequence of audio modality. 
It consists of two transformer~\cite{Vaswani2017_attuneed} stacks: audio transformers ${\rm TF}_{\rm A}$ and visual transformers ${\rm TF}_{\rm G}$.

Formally, we flatten the spatial dimensions of global visual feature $e^{\rm G}_t$, resulting in a feature vector $z^G_{1,t}\in\mathbb{R}^{H_{\textrm{G}}.W_{\textrm{G}}\times D_{\textrm{G}}}$. For a sequence of audio features $[e^{\rm A}_1, \cdots, e^{\rm A}_T]$, we apply average pooling across the height and width dimension, then stack along the temporal axis to obtain the feature vector $z^A_{1}\in\mathbb{R}^{T\times D_{\textrm{A}}}$. These feature vectors serve as the initial inputs for the AESPA module. To efficiently exchange information between the two modalities we utilize a set of learnable shared bottleneck tokens $b_{l, t}$.
The process inside $l$-th layer of AESPA can be formulated as follows: 
\begin{equation}
\begin{split}
&[z^{\mbox{\scriptsize A}}_{l+1,t},\hat{b}^{\mbox{\scriptsize A}}_{l+1,t}]=\mbox{TF}^{l}_{\mbox{\scriptsize A}}([ z^{\mbox{\scriptsize A}}_{l,t},b_{l,t}]),\\
&[z^{\mbox{\scriptsize G}}_{l+1,t},\hat{b}^{\mbox{\scriptsize G}}_{l+1,t}]=\mbox{TF}^{l}_{\mbox{\scriptsize G}}([ z^{\mbox{\scriptsize G}}_{l,t},b_{l,t}]),\\
&b_{l+1, t}= \mbox{AVG}(\hat{b}^{\mbox{\scriptsize A}}_{l+1},\hat{b}^{\mbox{\scriptsize G}}_{l+1}),
\end{split}
\end{equation}
where ${\rm TF}$ denotes transformer encoder layer. The bottleneck tokens are appended to the token sets of both modalities and processed together throughout $L$ layers of transformer stacks. Finally, we reshape the output of visual transformer $z^{\mbox{\scriptsize G}}_{L+1,t}$ to obtain the audio-enhanced visual feature $e^{\mbox{\scriptsize GA}}_{t}\in\mathbb{R}^{D_{\textrm{A}}\times H_{\textrm{G}}\times W_{\textrm{G}}}$.

The enhanced visual feature $e^{\mbox{\scriptsize GA}}_{t}$ is then fed into the patch extraction network $\pi$, which then produces the center coordinates ($\tilde{x}^t_c,\tilde{y}^t_c$) of an important patch $\tilde{v}_t$ on the image $V$.
These center coordinates are continuous values and are obtained as:
\begin{equation}
(\tilde{x}^t_c,\tilde{y}^t_c)=\pi\left(\{e^{\mbox{\scriptsize GA}}_1,\cdots,e^{\mbox{\scriptsize GA}}_t\}\right).
\end{equation}

Then, we obtain the coordinates of each pixel in the patch $(\tilde{x}^{t}_{ij},\tilde{y}^t_{ij})$ by adding a fixed offset $o_{ij}$ to $(\tilde{x}^t_c,\tilde{y}^t_c)$.
\begin{equation}
(\tilde{x}^t_{ij},\tilde{y}^t_{ij})= (\tilde{x}^t_c,\tilde{y}^t_c)+o_{ij}.
\end{equation}

As the corresponding coordinates $(\tilde{x}^{t}_{ij},\tilde{y}^t_{ij})$ have continuous values and need to be differentiable, we use bilinear interpolation~\cite{Wang2022_adafocusv2} to obtain the patch pixel value from the four pixels surrounding the coordinate.
Given the center coordinate of an important patch, we crop an original image to a patch of size $P \times P$.
We process this image patch with a heavier local visual network $f_L$, which is larger in parameter size to extract fine-grained features of the patch.
\begin{equation}
e^{\mbox{\scriptsize L}}_t=f_{\mbox{\scriptsize L}}(\tilde{v}_t).
\end{equation}
As the cropped patch is much smaller than the original image, the extraction process in network $f_{\mbox{\scriptsize L}}$ requires significantly less cost than processing the original image.

\subsection{Prediction Head}
Lastly, we build a classifier module based on the fusion of the extracted features. Before the classifier, we use an audio fusion transformer $\psi$ that has the same architecture as the standard transformer encoder~\cite{Vaswani2017_attuneed}.
It transforms the sequence of audio features $\{e^{\mbox{\scriptsize A}}_1, \cdots, e^{\mbox{\scriptsize A}}_T\}$, which are used as key and value, using the global visual feature $e^{\mbox{\scriptsize G}}_t$ as the query. The transformed audio feature at time $t$ is denoted as $e^{\textrm{AT}}_t$.
We concatenate the transformed audio feature with the  global feature $e^{\mbox{\scriptsize G}}$ and the local feature $e^{\mbox{\scriptsize L}}$.
We then feed the resulting feature into our classifier module $f^{\mbox{\scriptsize AV}}_C$.
\begin{equation}
p_t=f^{\mbox{\scriptsize AV}}_C\left(\{[e^{\mbox{\scriptsize G}}_1,e^{\mbox{\scriptsize L}}_1,e^{\textrm{AT}}_1], \cdots,[e^{\mbox{\scriptsize G}}_t,e^{\mbox{\scriptsize L}}_t,e^{\textrm{AT}}_t]\}\right).
\end{equation}
At each time step $t$, the classifier $f^{\mbox{\scriptsize AV}}_C$ generates a softmax prediction denoted as $p_t$. The classifier consists of fully-connected layers and aggregates features across the time steps using the \textit{max} operation.

\subsection{Training Techniques}
\paragraph{Video classification loss.} Firstly, we calculate the main loss of our network ($\mathcal{L}_{\mbox{\scriptsize p}}$), which is the cross-entropy loss of predicted output $p_t$ for all values of $t$.

\paragraph{Auxiliary visual loss.}
This loss aims to better train the visual modality encoders. We pass the global visual feature $e^{\mbox{\scriptsize{G}}}_t$ and the local visual feature $e^{\mbox{\scriptsize{L}}}_t$ to separate fully-connected layers $\mbox{FC}^{\mbox{\scriptsize{G}}}$ and $\mbox{FC}^{\mbox{\scriptsize{L}}}$, respectively.
Additionally, we use a classifier $f^{\rm V}_C$, which has the same structure as $f^{\rm AV}_C$, and takes as input the concatenated global and local visual features $e^{\mbox{\scriptsize{V}}}_t=[e^{\mbox{\scriptsize{G}}}_t,e^{\mbox{\scriptsize{L}}}_t]$ extracted up until time step $t$.
We calculate cross-entropy losses using these outputs.

\begin{equation} \label{eq1}
\begin{split}
\mathcal{L}_{\mbox{\scriptsize V}} = &
\frac{1}{T}\sum_{t=1}^{T}
\left(
\begin{aligned}
&L_{\mbox{\scriptsize CE}}(\mbox{FC}^{\mbox{\scriptsize{G}}}(e^{\mbox{\scriptsize{G}}}_t),y) +\\& L_{\mbox{\scriptsize CE}}(\mbox{FC}^{\mbox{\scriptsize{L}}}(e^{\mbox{\scriptsize{L}}}_t),y) +\\
 & L_{\mbox{\scriptsize CE}}\left(f_C^{\mbox{\scriptsize{V}}}(\{e^{\mbox{\scriptsize V}}_1, \cdots,e^{\mbox{\scriptsize V}}_t\}\right),y)
\end{aligned}
\right).
\end{split}
\end{equation}

\paragraph{Auxiliary audio loss.}
Similarly, we also apply auxiliary loss on the audio modality. We use FC classifier $\mbox{FC}^{\mbox{\scriptsize{A}}}$ and audio sequence classifier $f_C^{\mbox{\scriptsize{A}}}$ then calculate the losses as,
\begin{equation} \label{eq2}
\mathcal{L}_{\mbox{\scriptsize A}} = \frac{1}{T}
\sum_{t=1}^{T}
\left(
\begin{aligned}
&L_{\mbox{\scriptsize CE}}(\mbox{FC}^{\mbox{\scriptsize{A}}}(e^{\mbox{\scriptsize{A}}}_t),y) +\\
&L_{\mbox{\scriptsize CE}}(f_C^{\mbox{\scriptsize{A}}}\left(\{e^{\mbox{\scriptsize A}}_1, \cdots,e^{\mbox{\scriptsize A}}_t\}\right),y)
\end{aligned}
\right).
\end{equation}

\paragraph{Masked visual token reconstruction.}
To enhance the robustness of the AV-TeST, we drop a subset of global visual tokens $\{\tilde{e}^{\rm G}_{T_{G+1}}, \cdots , \tilde{e}^{\rm G}_{T}\}$, and train the model to recover the missing part using the remaining features.
We first embed the remaining features, and concatenate audio embeddings $\{\tilde{e}^{\rm A}_{T_{\rm G+1}}, \cdots , \tilde{e}^{\rm A}_{T}\}$ with mask tokens $\mathcal{M}$ to match the input dimension.
These are then passed to the transformer encoders, producing $\{{e}^{\rm TF}_{1}, \cdots , \tilde{e}^{\rm TF}_{T}\}$.
Next, the partial outputs $\{{e}^{\rm TF}_{\rm G+1}, \cdots , \tilde{e}^{\rm TF}_{T}\}$ are fed to a mask encoder to reconstruct the missing global visual tokens, generating $\{\hat{e}^{\rm G}_{T_{\rm G+1}}, \cdots , \hat{e}^{\rm G}_{T}\}$.
We calculate the L2 loss between the embedded visual tokens $\{\tilde{e}^{G}_{T_{\rm G+1}}, \cdots, \tilde{e}^{\rm G}_{T}\}$ and the reconstructed visual tokens $\{\hat{e}^{\rm G}_{T_{\rm G+1}}, \cdots , \hat{e}^{\rm G}_{T}\}$ as $\mathcal{L}_{\mbox{\scriptsize{mask}}}$ (see Fig.~\ref{fig:AVTST_train}). This loss enables AV-TeST to work robustly even with limited numbers of global visual tokens at the inference stage.

\paragraph{Saliency loss.}
In order to train the AV-TeST ($\mbox{TF}_{\mbox{\scriptsize AV}}$) to produce the saliency score without any ground truth frame importance, we generate pseudo labels $\tilde{s}_t$.
We first obtain softmax predictions $p^{'}_t$ using the classifier $f^{\mbox{\scriptsize{AV}}}_C$ with only the features at time step $t$, without using  features from previous time steps or the \textit{max} operation for feature aggregation.
\begin{equation}
p^{'}_t = f^{\mbox{\scriptsize{AV}}}_C \left([e^{\mbox{\scriptsize G}}_t,e^{\mbox{\scriptsize L}}_t,e^{\textrm{AT}}_t]\right)
\end{equation}
We then obtain a confidence score of each frame by normalizing $p^{'}_t$ with the maximum prediction logit across the classes and across the time steps.
We minimize the difference between the predicted saliency scores $s_t$ and confidence score $\tilde{s}_t$ with L1 loss (see Fig.~\ref{fig:AVTST_train}).
\begin{equation}
\mathcal{L}_{s} = L_1(s_t, \tilde{s}_t), \;\;\;\tilde{s}_t=\dfrac{\max_c {p}^{'}_{c,t}}{\max_t \max_c {p}^{'}_{c,t}},
\end{equation}
where $p^{'}_{c,t}$ denotes the prediction logit value of class $c$ at time step $t$. The denominator normalizes the pseudo labels, ensuring that the maximum value across all time steps and classes is 1.

\paragraph{Ordered AV logits loss.} 
We reorder $e^{\mbox{\scriptsize AV}}$ based on the saliency scores ($\textbf{s}$), followed by simulating the limited frame inference by applying Gumbell-Softmax sampling~\cite{jang2016categorical} on the saliency scores, which selects only a few frames from the entire sequence. We then compute the cross-entropy loss for the video classification using the reordered sequence.
\begin{equation} \label{eq3}
\begin{split}
\mathcal{L}_{\mbox{\scriptsize ord}} = L_{\mbox{\scriptsize CE}}(f^{\mbox{\scriptsize AV}}_C\left(\{e^{\mbox{\scriptsize AV}}_{i_1},\cdots, e^{\mbox{\scriptsize AV}}_{i_T}\}\right),y),
\end{split}
\end{equation}
where $i_t$ represents the temporal index of the frame with the $t$-th highest saliency score that is selected after the sampling.

We train AVGN by summing all the loss functions ($\mathcal{L}_{\mbox{\scriptsize p}}, \mathcal{L}_{\mbox{\scriptsize V}}, \mathcal{L}_{\mbox{\scriptsize A}}, \mathcal{L}_{s}, \mathcal{L}_{\mbox{\scriptsize{mask}}}, \mathcal{L}_{\mbox{\scriptsize ord}}$).
To stabilize the training, we stop the gradient from $\pi$ and $\mbox{TF}_{\mbox{\scriptsize AV}}$ to $f_G$ and $f_A$, and from $\psi$ to $f_G$.
\subsection{Inference}
At the initial temporal glance stage, we can achieve efficiency by limiting global visual feature extractions up to only $T_g$ frames, and instead, put mask tokens $\mathcal{M}$ to replace the unextracted features.
\begin{equation}
    e^{\mbox{\scriptsize {AV}}}_t = \begin{cases}
         \mbox[e^{\mbox{\scriptsize {G}}}_t,e^{\mbox{\scriptsize{A}}}_t] & \mbox{if} \ \ t\leq T_g, \\
         [\mathcal{M},e^{\mbox{\scriptsize{A}}}_t] & \mbox{otherwise}. \\
    \end{cases}
\end{equation}
Then, for the spatial glance and the subsequent process we can process only $k$ frames based on their saliency scores.

%% file: 04_experiments.tex
\input{tabs/sota_comparison}
\section{Experiments}
\subsection{Setup}
\paragraph{Datasets.} We used three large-scale video recognition datasets, \ie, ActivityNet~\cite{caba2015_activitynet}, FCVID~\cite{Jiang2018_FCVID}, and Mini-Kinetics~\cite{kay2017_kinetics}, and adopted their official training-validation splits. Following the common practice~\cite{Gao2020_ListenToLook, Wang2019_TSN, Meng2020_AR-net, Meng2021_AdaFuse,Wang2021_adafocus,Wang2022_adafocusv2,Zheng2020_DynamicSamplingNetworks}, we evaluated the performance of different methods via Top-1 accuracy (Top-1 Acc.) for Mini-Kinectics and mean average precision (mAP) for the other datasets.

\paragraph{Implementation details.} For every video, we temporally sampled 16 pairs of image frames and audio clips. In the training stage, the frames are firstly grouped into 16 bins, and then one sample is randomly taken out from each bin. Then, we set the order of the frames with strategy following \cite{Ghodrati2021_frameexit}. For the audio clips, we convert them to 1-channel audio-spectrograms of size $96 \times 64$ (960 msecs. and 64 frequency bins). We followed ~\cite{Wang2021_adafocus,Wang2022_adafocusv2} for the visual data pre-processing and augmentation, and performed time masking and frequency masking on the audio. For inference, we resized all frames to $256 \times 256$ and performed center-crop with size of $224 \times 224$. The value of $T_G$ is set to $12$ unless stated otherwise.

\subsection{Comparison with state-of-the-art}
In Table~\ref{tab:main_result_table}, we compared AVGN with several competitive action recognition baselines~\cite{Wu2019_LiteEval, Korbar2019_SCSampler, Gao2020_ListenToLook,Meng2020_AR-net,Wu2022_AdaFrame,Meng2021_AdaFuse,Sun2021_VideoIQ,Wang2022_adafocusv2,Wang2022_AdaFocusv3} on  multiple datasets: ActivityNet~\cite{caba2015_activitynet}, FCVID~\cite{Jiang2018_FCVID}, and Mini Kinetics~\cite{kay2017_kinetics}.
We evaluated the model based on the mAP and computational cost measured in terms of floating point operations per second (FLOPS) averaged over the entire videos during the inference stage.
We report the performance of AVGN in three settings: (i) AVGN with the maximum number of frames allowed for the prediction set to 14 ($k = 14$); (ii) AVGN with $T_G=4$ and $k=5$ for ActivityNet and FCVID, $k=3$ for Mini-Kinetics; (iii) AVGN with an early exit~\cite{Ghodrati2021_frameexit}, with the number of glances set to 2 ($T_{G}=2$) for cost-effective inference. 

We see that AVGN outperformed ListenToLook~\cite{Gao2020_ListenToLook}, which also utilizes the audio modality in its ImgAud backbones, with an mAP gain of 3.7\% on ActivityNet using less than a third of GFLOPs.
Additionally, AVGN performs competitively with AdaFocus variants and outperforms both in trimmed video recognition with better efficiency on Mini-Kinetics (73.1\% mAP with 6.1 GFLOPs vs. 72.9\% mAP with 8.6 GFLOPs on AdaFocusV3~\cite{Wang2022_AdaFocusv3}). AVGN achieved 1.2\% mAP gain over AdaFocusV2~\cite{Wang2022_adafocusv2} while requiring 2.6 fewer GFLOPs on ActivityNet. On Mini-Kinetics, AVGN outperformed AdaFocusV2 by 2.5\% accuracy with 2.6 fewer GFLOPs. Focusing only on the mAP metric, AVGN achieves the highest scores in ActivityNet and Mini-Kinetics. From the results of setting (ii) and (iii), we see the reliable performance of AVGN for low-cost inference, either by simple method of limiting the $k$ or threshold method as \cite{Ghodrati2021_frameexit}. AVGN$^+$ also shows competitive results with significantly fewer GFLOPs than other methods. For example, it achieves comparable results with OCSampler~\cite{Lin2022_OCSampler} with half the GFLOPs on the FCVID dataset. It is worth noting that the slightly lower performance of AVGN on FCVID can be attributed to the fact that FCVID is a large-scale dataset that contains diverse video genres, whereas ActivityNet and Mini-Kinetics focus more on human action. Nonetheless, our experiments demonstrate that AVGN provides an efficient solution for video recognition tasks, surpassing existing state-of-the-art methods in terms of both accuracy and computational cost.

\begin{figure}[t]
\centering
    \includegraphics[width=\columnwidth]{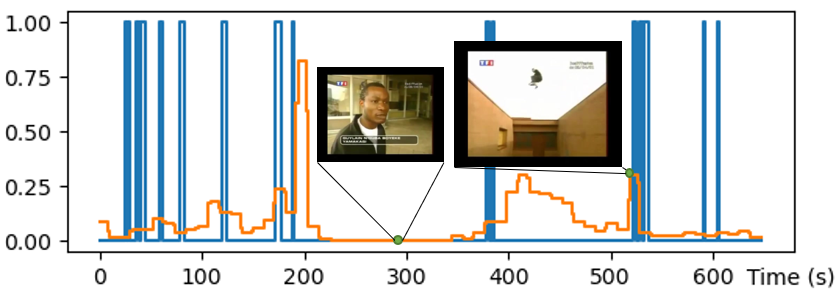}
    \vspace{-0.3in}
    \caption{\textbf{Qualitative result of highlight detection on a long video.} We show {\bf\textcolor{orange}{predicted}} important and unimportant frames that correlate with the {\bf\textcolor{skyblue}{ground truth}} for the category \texttt{parkour}.}  
    \label{fig:HD_qual_result}
    \vspace{-0.15in}
\end{figure}

\begin{table}[t]
  \centering
  \begin{footnotesize}
  \setlength{\tabcolsep}{2mm}{
  \renewcommand\arraystretch{1.1}
  \resizebox{\columnwidth}{!}{
  \begin{tabular}{c|cccccc}
  Method & VESD & LM-A & LM-S & MN & CHD & {AVGN} \\
  \hline
  Property & WS,ED & WS,ED & WS,ED & WS,ED & US & {ZS, ED} \\
  mAP & 0.423 & 0.524 & 0.563 & 0.698 & 0.527 & {0.557} \\
  \end{tabular}}}
  \vspace{-0.1in}
  \caption{\textbf{Highlight detection results on the TVSum Dataset.} 
  We report the top-5 mAP results along with their properties (WS = Weakly Spv., US = Unspv., ZS = Zero-shot, ED = External Data).}
   \label{tab:sota_hd}
  \end{footnotesize}
  \vspace{-0.2in}
\end{table}

\input{tabs/k_glances}

\input{tabs/ablation_comp}

\subsection{Long Videos Understanding}
To demonstrate the effectiveness of AVGN on long videos, we use the ActivityNet-trained AVGN for the task of zero-shot Highlight Detection (HD). We use the TVSum\cite{song2015_tvsum} dataset, consisting of 2 to 10 minutes videos, with 4.2 minutes on average. The results in Table \ref{tab:sota_hd} and Figure \ref{fig:HD_qual_result} show AVGN performs competitively to models dedicatedly trained for HD, despite being trained for action recognition on another dataset. This highlights AVGN's effectiveness in understanding long videos across datasets and tasks.

\subsection{Ablation Studies}

\paragraph{Effect of the number of glanced frames.}
To investigate the effect of the number of glances, we conducted experiments with varying numbers of glanced frames ($T_{G}$) and the maximum number of image-audio pairs ($k$) passed to the spatial glance stage. Table~\ref{tab:k_glances} presents the results.
As shown in the table, increasing the number of glanced frames ($T_{G}$) leads to higher mAP due to more precise saliency score estimations. However, this comes at the expense of higher GFLOPs, indicating a trade-off between computational cost and accuracy.
For instance, at $k=6$, increasing $T_G$ from 2 to 16 results in a mAP gain, from 75.6\% to 78.3\%, but at the same time, the GFLOPs value increases from 10.9 to 14.1. 
When all frames are passed through ($k=16$) so that there is no temporal filtering, the difference in performance between different $T_G$ values is minuscule.
The results show that depending on the $k$, a moderate increase in $T_G$ can result in significant gain in mAP, but beyond a certain point, the improvement in performance becomes marginal.

\begin{table}[t]
  \centering
  \begin{footnotesize}
  \setlength{\tabcolsep}{1mm}{
  \renewcommand\arraystretch{1.05}
  \resizebox{\columnwidth}{!}{
  \begin{tabular}{c|cccccc}
  \multirow{2}{*}{\parbox{1.2cm}{Temporal Sampling}} & \multicolumn{2}{c}{$k=4$} & \multicolumn{2}{c}{$k=8$} & \multicolumn{2}{c}{$k=12$} \\
  & mAP & GFs & mAP & GFs & mAP & GFs\\
  \hline
  Uniform &  67.3\% & 7.4 & 76.6\% & 14.2 & 78.8\% & 20.9   \\
  AV-TeST($T_G$=12) & 75.4\% & 10.5 & 78.8\% & 16.0 & 79.9\% & 22.0 \\
  \end{tabular}}}
  \vspace{-0.13in}
  \caption{\textbf{Ablation study on temporal glance.} }
   \label{tab:abl_temporal}
  \end{footnotesize}
  \vspace{-0.13in}
\end{table}

\begin{table}[t]
  \centering
  \begin{footnotesize}
  \setlength{\tabcolsep}{1mm}{
  \renewcommand\arraystretch{1.05}
  \resizebox{\columnwidth}{!}{
  \begin{tabular}{c|ccccc}
  Spatial Sampling & $k=1$ & $k=4$ & $k=8$ & $k=12$ & $k=16$ \\
  \hline
  Center Crop & 44.2\% & 59.3\% & 67.6\% & 70.0\% & 71.4\%  \\
  Patch Network ($\pi$) & 48.0\% & 62.2\% & 69.8\% & 71.9\% & 73.6\%   \\
  AESPA + $\pi$ & 48.6\% & 62.9\% & 70.4\% & 72.8\% & 74.1\%  \\
  \end{tabular}}}
  \vspace{-0.13in}
  \caption{\textbf{Ablation study on spatial glance.} }
   \label{tab:abl_spatial}
  \end{footnotesize}
  \vspace{-0.1in}
\end{table}
\input{tabs/ablation_loss}

\paragraph{Contribution of AVGN components.}
Table~\ref{tab:abl_study} presents the results of ablation studies on the AVGN model.
We first compared the performance of the model with and without AV-TeST (in Exp. 2 vs. Exp 1 and Exp. 3 vs. Exp. 4).
The results show that AV-TeST improves the performance even when the number of frames for inference is limited and irrespective of whether is available of not.
Exp. 1 vs. Exp. 3 demonstrates that incorporating audio modality can increase the mAP by around 2\%. 
Exp. 4 vs. Exp. 5 shows that AESPA module boosts the performance about 1\%.
Exp. 5 vs. Exp. 6 indicates that $\psi$ has a significant impact on the performance, especially when $k$ is low, \eg, 4.6\% mAP gain when $k=1$. 
Comparing Exp. 3 to 6, the performance improves progressively with the addition of more components, especially when the audio modality is fully utilized.
The model achieves the highest accuracy of 80.2\% in Exp. 6 when all components are used with $k=16$.
Comparing AV-TeST sampling with uniform temporal sampling in Table 
~\ref{tab:abl_temporal}, AVGN achieves better mAP while not adding much GFLOPs (GFs). To assess the effectiveness of spatial glancing, we studied the predictive power of the extracted visual patch in the experiment presented in Table \ref{tab:abl_spatial}. We use local features ($e^L_t$) that have been aggregated through the temporal axis using a linear layer classifier for prediction.  Improvements are achieved with the extraction patch network, especially when combined with our AESPA for audio.

\begin{figure}[t]
\centering
    \includegraphics[width=0.89\columnwidth]{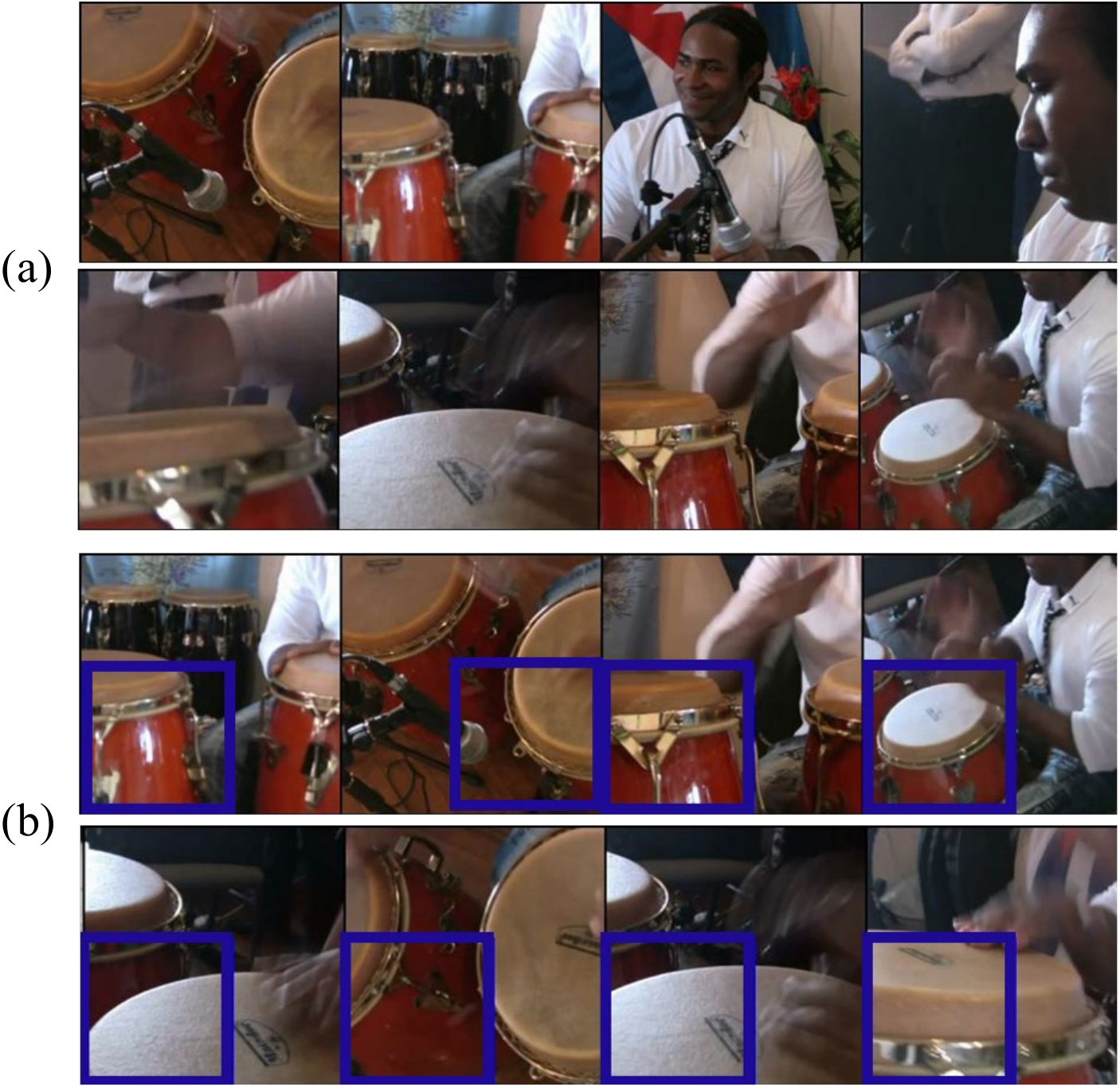}
    \vskip -0.1in
    \caption{\textbf{Qualitative result.} Part \textbf{(a)} shows the original first 8 frames of the video and part \textbf{(b)} shows the Top-8 salient frames. The images are ordered from left to right and the bounding box on each frame highlights the important spatial area.}
    \label{fig:qual_result}
    \vskip -0.2in
\end{figure}

\paragraph{Ablation study of losses.}
Table~\ref{tab:abl_loss} shows the ablation experiments of training losses.
First, we confirm the importance of the modality auxiliary losses, specifically the visual $\mathcal{L}_{\mbox{\scriptsize{V}}}$ and audio $\mathcal{L}_{\mbox{\scriptsize{A}}}$ losses.
Removing either of these losses led to significant drop in mAP.
The masking loss $\mathcal{L}_{\rm mask}$ significantly improved the performance, especially when processing a lower number of $k$, \eg, 2.7\% mAP gain when $k=1$.
Lastly, the ordered AV logits loss $\mathcal{L}_{ord}$ contributed to a slight yet noteworthy improvement to the performance.

\subsection{Qualitative result}
We present the qualitative result of our model in the form of salient frames and their important spatial patches extracted from a video. In Fig.~\ref{fig:qual_result} (a), we show the first 8 frames directly sampled from the video, and in Fig.~\ref{fig:qual_result} (b), we display 8 highest saliency score frames, with bounding boxes that highlight the important patches.
The results demonstrate that our AVGN effectively identified the important part of the ``\texttt{playing conga}'' action class, which is the musical instrument. From a temporal perspective, the model prioritized frames where the conga is clearly visible.

%% file: tabs/sota_comparison.tex
\begin{table*}[t]
  \centering
  \footnotesize
  \setlength{\tabcolsep}{2mm}{
  \renewcommand\arraystretch{1}
  \resizebox{0.9\textwidth}{!}{
  \begin{tabular}{c|c|c|cccccc}
  \multirow{2}{*}{Methods} & \multirow{2}{*}{Published on}& \multirow{2}{*}{Backbones}  & \multicolumn{2}{c}{ActivityNet} &  \multicolumn{2}{c}{FCVID} & \multicolumn{2}{c}{Mini-Kinetics}  \\
  &&& mAP{\up} & GFLOPs{\down} & mAP{\up} &  GFLOPs{\down} & Acc.{\up} &  GFLOPs{\down} \\
  \midrule
  LiteEval \cite{Wu2019_LiteEval} & {NeurIPS'19} &MN2+RN  & 72.7\% & 95.1 & 80.0\% & 94.3  & 61.0\% & 99.0 \\
  SCSampler \cite{Korbar2019_SCSampler} & {ICCV'19} &MN2+RN & 72.9\%  & 42.0 & 81.0\% & 42.0  & 70.8\%  & 42.0 \\
  ListenToLook \cite{Gao2020_ListenToLook} & {CVPR'20} &MN2+RN & 72.3\%  & 81.4 & -- & --  & -- & -- \\ 
  ListenToLook \cite{Gao2020_ListenToLook} & {CVPR'20} &IA & 76.4\%  & 76.1 & -- & --  & -- & --  \\
  AR-Net \cite{Meng2020_AR-net} & {ECCV'20} &MN2+RN & 73.8\% & 33.5 & 81.3\% & 35.1  & 71.7\% & 32.0  \\
  AdaFrame \cite{Wu2022_AdaFrame} & {T-PAMI'21} &MN2+RN & 71.5\% & 79.0 & 80.2\% & 75.1  & -- & --  \\
  VideoIQ \cite{Sun2021_VideoIQ} & {ICCV'21} &MN2+RN & 74.8\% &  28.1 & {82.7\%} & {27.0}  & 72.3\% &  20.4 \\
  OCSampler \cite{Lin2022_OCSampler} & {CVPR'22} &MN2$^\dagger$+RN & 76.9\% &  21.7 & 82.7\% & 26.8  & 72.9\% &  17.5 \\
  AdaFocusV2 \cite{Wang2022_adafocusv2} & {CVPR'22} &MN2+RN & 79.0\% &  27.0 & 
  \underline{85.0\%} & 27.0  & \underline{75.4\%} & 27.0 \\
  AdaFocusV2$^+$\cite{Wang2022_adafocusv2} & {CVPR'22} &MN2+RN & 74.8\% &  9.9 & 82.7\% & \underline{10.1} & 72.3\% & \underline{6.3} \\
  AdaFocusV3 \cite{Wang2022_AdaFocusv3} & {ECCV'22} &MN2+RN & \underline{79.5\%} &  21.7 & 
  \textbf{85.9\%} & 26.8  & 75.0\% & 17.5 \\
  AdaFocusV3$^+$\cite{Wang2022_AdaFocusv3} & {ECCV'22} &MN2+RN & 76.9\% &  10.9 & 82.7\% & \textbf{7.8} & 72.9\% & 8.6 \\
  \rowcolor{Gray}
  AVGN ($k$=14)  & -- &MN2+RN  & \ \textbf{80.2\%}
    & 24.4 & 84.1\% & 24.4 & \textbf{77.9\%} & 25.0  \\
    \rowcolor{Gray}
AVGN$^{*}$ ($T_G=4$)& -- &MN2+RN  & 74.7\% & \textbf{9.6}
  & 82.2\% & 10.4  &  72.5\% &  6.4\\
  \rowcolor{Gray}
AVGN$^+$ ($T_G=2$)& -- &MN2+RN  & 75.8\% & \underline{9.7}
  & 82.4\% & 10.2  &  73.1\% &  \textbf{6.1}\\
  \end{tabular}}}
  \vspace{-0.1in}
  \caption{\textbf{Comparison of AVGN and baselines on three benchmark datasets}. MN2,IA,RN denotes MobileNet-V2,ImgAud\cite{Gao2020_ListenToLook}, and ResNet respectively. The best two results are \textbf{bold-faced} and \underline{underlined}, respectively. GFLOPs refer to the average computational cost for processing every videos. For AVGN$^*$, we use $k=5$ for ActivityNet and FCVID, $k=3$ for Mini-Kinetics. $\dagger$ represents the addition of TSM~\cite{Lin2019_TSMVideoUnderstanding} and $^+$ represents the addition of early exit~\cite{Ghodrati2021_frameexit}.
  {\color{plus}{$\uparrow$}: \rm The higher the better.}
  {\color{minus}{$\downarrow$}: \rm The lower the better.}
    \label{tab:main_result_table}
  }
  \vskip -0.15in
\end{table*}

%% file: tabs/k_glances.tex
\begin{table}[t]
  \centering
  \footnotesize
  \setlength{\tabcolsep}{1mm}{
  \resizebox{\columnwidth}{!}{
  \begin{tabular}{c|cccccc}
  \multirow{2}{*}{\shortstack{Number of\\glances ($T_{G}$)}} & \multicolumn{2}{c}{$k$=1} & \multicolumn{2}{c}{$k$=6}  & \multicolumn{2}{c}{$k$=16}\\
& mAP & GFLOPs & mAP & GFLOPs & mAP & GFLOPs \\
\midrule
2 & 55.3\% & 2.7 & 75.6\% & 10.9 & \textbf{80.2}\% & 27.6 \\
4 & 56.4\% & 3.3 & 75.9\% & 11.1 & 80.1\% & 27.6 \\
8 & 57.5\% & 4.7 & 76.8\% & 11.6 & 80.0\% & 27.6 \\
12 & \textbf{57.7}\% & 6.0 & 77.8\% & 12.8 & \textbf{80.2}\% & 27.6 \\
16 & 57.6\% & 7.3 & \textbf{78.3}\% & 14.1 & \textbf{80.2}\% & 27.6 \\
  \end{tabular}}}
  \vspace{-0.1in}
  \caption{\textbf{Effect of the number of glances} ($T_{G}$) on performance (mAP) and computation (FLOPs).
  Experiments are conducted on ActivityNet~\cite{caba2015_activitynet} and $k$ denotes the maximum number of image-audio pairs passed to the spatial glance stage.}
    \label{tab:k_glances}
  \vskip -0.2in
\end{table}


%% file: tabs/ablation_comp.tex
\begin{table*}[h]
  \centering
  \footnotesize
  \setlength{\tabcolsep}{3mm}{
  \renewcommand\arraystretch{0.95}
  \resizebox{0.9\textwidth}{!}{
  \begin{tabular}{c|cccc|cccccc}
  Exp.&Audio&AV-TeST&AESPA&$\psi$& $k$=1 & $k$=4 & $k$=6 & $k$=8 & $k$=12 & $k$=16 \\
  \midrule
  1 & {\xmark} & {\xmark} & {\xmark} & {\xmark} &\ 46.6\% & 63.6\% & 69.0\% & 72.7\% & 75.6\% & 76.9\%  \\
  2 & {\xmark} & {\cmark} & {\xmark} & {\xmark} &\ 50.1\% & 70.4\% & 73.0\% & 74.5\% & 76.0\% & 76.9\%  \\
  3 & {\cmark} & {\xmark} & {\xmark} & {\xmark} &\ 48.6\% & 65.2\% & 70.9\% & 74.4\% & 77.1\% & 78.4\%  \\
  4 & {\cmark} & {\cmark} & {\xmark} & {\xmark} &\ 51.5\% & 70.3\% & 73.9\% & 76.0\% & 78.1\% & 78.4\%  \\
  5 & {\cmark} & {\cmark} & {\cmark} & {\xmark} &\ 52.1\%
  & 71.5\% & 75.1\% & 76.7\% & 79.0\% & 79.7\% 
  \\
  6 & {\cmark} & {\cmark}& {\cmark} & {\cmark}&  \ 
  \textbf{57.7\%} & \textbf{75.4\%} & \textbf{77.8\%} & \textbf{78.8\%} & \textbf{79.9\%} & \textbf{80.2\%}  \\
  \end{tabular}}}
  \vspace{-0.1in}
  \caption{\textbf{Ablation study on AVGN key components}.
  We report the mAP results on ActivityNet~\cite{caba2015_activitynet}.
  We alternately add or remove audio modality, AV-TeST transformer (${\rm TF_ {AV}}$), AESPA module ($æ$), and the audio transformation network ($\psi$) at the classifier.
  }
  \vspace{-0.2in}
\label{tab:abl_study}
\end{table*}

%% file: tabs/ablation_loss.tex
\begin{table}[h]
  \centering
  \begin{footnotesize}
  \setlength{\tabcolsep}{2mm}{
  \renewcommand\arraystretch{1}
  \resizebox{\columnwidth}{!}{
  \begin{tabular}{cccc|cccc}
  $\mathcal{L}_{\mbox{\scriptsize V}}$ & $\mathcal{L}_{\mbox{\scriptsize U}}$ & $\mathcal{L}_{\mbox{\scriptsize{mask}}}$ & $\mathcal{L}_{ord}$ & $k=1$ & $k=4$ & $k=8$ & $k=16$ \\
  \midrule
  \textcolor{black}{\xmark} & \textcolor{black}{\cmark} & \textcolor{black}{\cmark} & \textcolor{black}{\cmark} &\ 56.9\% & 73.0\% & 76.5\% & 78.4\%   \\
  \textcolor{black}{\cmark} & \textcolor{black}{\xmark} & \textcolor{black}{\cmark} & \textcolor{black}{\cmark} &\ 54.6\% & 73.0\% & 77.2\% & 78.6\%   \\
  \textcolor{black}{\cmark} & \textcolor{black}{\cmark} & \textcolor{black}{\xmark} & \textcolor{black}{\cmark} &\ 55.0\% & 73.7\% & 77.6\% & 79.5\%   \\
  \textcolor{black}{\cmark} & \textcolor{black}{\cmark} & \textcolor{black}{\cmark} & \textcolor{black}{\xmark} &\ 57.0\% & 75.2\% & 78.8\% & 80.2\%  \\
  \textcolor{black}{\cmark}& \textcolor{black}{\cmark} & \textcolor{black}{\cmark} & \textcolor{black}{\cmark}&  \ \textbf{57.7\%} & \textbf{75.4\%} & \textbf{78.8\%} & \textbf{80.2\%}
  \end{tabular}}}
  \vspace{-0.1in}
  \caption{\textbf{Ablation study on training losses.} 
  We report the mAP results on ActivityNet~\cite{caba2015_activitynet}.}
   \label{tab:abl_loss}
  \end{footnotesize}
  \vspace{-0.2in}
\end{table}

%% file: 05_conclusion.tex
\section{Conclusion}
We have proposed an efficient video recognition network called Audio-Visual Glance Network (AVGN) that selectively processes spatiotemporally important parts of a video.
To improve network efficiency, we incorporated a cost-effective audio modality in addition to the visual modality. AVGN utilizes AV-TeST, a multimodal transformer that estimates salient frame saliency to achieve temporal efficiency. For spatial efficiency, we combine a recurrent patch extraction network and Audio Enhanced Spatial Patch Attention (AESPA) module to find important spatial patch using audio-enhanced coarse visual features. Our experiments have shown that AVGN performs competitively among state-of-the-art methods. Moreover, our ablation studies indicate that all of our model building blocks work collaboratively to contribute to model achievement.

%% file: 06_appendix.tex
\appendix
\label{sec:appendix}
\maketitle

\section{Notation List}
For the convenience of the reader, we listed the Table of Notation containing frequently used notations along with their definition in Table \ref{tab:notation_list}.
\section{Datasets}
\paragraph{ActivityNet-1.3} \cite{caba2015_activitynet} contains 10,024 training videos and 4,926 validation videos sorted into 200 human action categories. The average duration is 117 seconds.
\paragraph{FCVID} \cite{Jiang2018_FCVID} contains 45,611 videos for training and 45,612 videos for validation, which are annotated into 239 classes. The average duration is 167 seconds.
\paragraph{Mini-Kinetics} is a subset of the Kinetics \cite{kay2017_kinetics} dataset. We establish it following \cite{Wu2019_LiteEval, Meng2020_AR-net, Meng2021_AdaFuse, Ghodrati2021_frameexit}. The dataset include 200 classes of videos, 121k for training and 10k for validation. The average duration is around 10 seconds~\cite{kay2017_kinetics}.
\section{Implementation Detail}
\subsection{Network Architecture}
\paragraph{Encoders.} For audio encoder $f_{\mbox{\scriptsize{A}}}$ and $f_{\mbox{\scriptsize{G}}}$ we use MobileNetV2 \cite{Sandler2018_MobileNetv2} and for local visual encoder $f_{\mbox{\scriptsize{L}}}$ we use ResNet-50 \cite{He2016_ResNet}. We use a patch size of $128 \times 128$ for the input to $f_L$, thus the size of the patch extracted by the patch extraction network is also the same. To encode a single image, the $f_{\mbox{\scriptsize{G}}}$ requires 0.33 GFLOPs and $f_{\mbox{\scriptsize{L}}}$ requires 1.35 GFLOPs, meanwhile to encode the whole audio sequence $f_{\mbox{\scriptsize{A}}}$ requires 0.68 GFLOPs.
\paragraph{AV-TeST.} In our implementation we construct $\textnormal{TF}_{\textnormal{AV}}$ using a multi-head attention transformer \cite{Vaswani2017_attuneed} with $256$ encoder dimension size, $2$ stacks, and $4$ heads. As the input to the transformer is concatenated audio-visual feature, for each modality we embed them to $128-d$ vectors with separate linear embedding layers. To reconstruct the visual token, we utilize a transformer with the same architecture and append a linear embedding layer at the end.
\paragraph{AESPA.} In our implementation of AESPA module, we use the same  transformer architecture for both audio and visual modality. To minimize the computational burden, we reduce the incoming channel of both audio and visual modality to $256$. Then we use the reduced feature maps as input to the bottleneck fusion transformers. Each modality transformer consists of $4$ stack of encoder with $4$ heads. We use $4$ bottleneck tokens to be appended to the modality tokens.

\paragraph{Training Details.}
To train the network, we use an SGD optimizer with cosine learning rate annealing and a momentum of 0.9. The L2 regularization co-efficient is set to 1e-4. The two encoders $f_{\textnormal{G}}$ and $f_{\textnormal{L}}$ are initialized using the ImageNet pre-trained models\footnote{In most cases, we use the 224x224 ImageNet pre-trained models provided by PyTorch \cite{paszke2019_pytorch}.}, while the rest of the network is trained from random initialization. The size of the mini-batch is set to 24. The initial learning rates of $f_{\textnormal{G}}$, $f_{\textnormal{A}}$, $f_{\textnormal{L}}$, $f_{\textnormal{C}}$, $\pi$, $\ae$, and $\textnormal{TF}_{\textnormal{AV}}$ are set to 0.001, 0.001, 0.002, 0.01, 2e-4, 2e-4, and 0.01. We use a masking ratio of $0.75$ for $L_{\textnormal mask}$, and for Gumbell-Softmax we use $5$ as the temperature value
\begin{table*}[h]
    \centering
    \renewcommand\arraystretch{1.2}
    \resizebox{\textwidth}{!}{\begin{tabular}{ll|ll}
        \toprule
        \textbf{Variables} &  & \multicolumn{2}{l}{\textbf{Functions}} \\
        Symbol & Definition & Symbol & Definition \\ 
        $t$ & Frame or time index & $f_A$ & Audio encoder\\
        $a_t$ & Audio spectrogram clip at time & $f_G$ & Global visual encoder\\
        $v_t$ & Input image frame at time $t$ & $f_L$ & Local visual encoder\\
        $y$ & label class &  $\mbox{TF}_{\scriptsize{\mbox{AV}}}$ & AV-TeST Transformer Network\\
        $e^{\mbox{\scriptsize{A}}}_t$ & Audio feature at time $t$ &  $\mbox{FC}_s$ & Saliency score prediction head\\
        $e^{\mbox{\scriptsize{G}}}_t$ & Coarse/Global visual feature at time $t$ & $\ae$ & Audio Enhanced Spatial Patch Attention (AESPA) module\\
        $z^{\mbox{\scriptsize{A}}}_{l,t}$ & AESPA audio vector at layer $l$ at time $t$ & $\mbox{TF}^l_{\mbox{\scriptsize{A}}}$ & AESPA audio transformer at layer $l$ \\
        $z^{\mbox{\scriptsize{G}}}_{l,t}$ & AESPA visual vector at layer $l$ at time $t$  & $\mbox{TF}^l_{\mbox{\scriptsize{G}}}$ & AESPA visual transformer at layer $l$ \\
        $e^{\mbox{\scriptsize{GA}}}_t$ & Enhanced Coarse/Global visual feature at time $t$  & $\pi$ & Spatial patch extraction network\\
        $e^{\mbox{\scriptsize{L}}}_t$ & Fine/Local visual feature at time $t$  & $\psi$ & Fusion transformer\\
        $e^{\mbox{\scriptsize{TF}}}_t$ & Audio-visual feature for AV-TeST input $t$  & $f^{\mbox{\scriptsize{AV}}}_C$ & Audio-visual classifier\\
        $s_t$ & Frame saliency score at time $t$ & $f^A_C$ & Auxiliary audio prediction head\\ 
        $\tilde{e}^{\mbox{\scriptsize{A}}}_t$ & Transformed audio feature at time $t$& $\mbox{FC}^{\mbox{\scriptsize{G}}}$  & Auxiliary frame-wise global visual prediction head\\
        $(\tilde{x}^t_c,\tilde{y}^t_c)$ & Center coordinates $t$ & $\mbox{FC}^{\mbox{\scriptsize{L}}}$ & Auxiliary frame-wise local visual prediction head\\
        $\tilde{v}_t$ & Visual patch at time $t$ & $\mbox{FC}^{\mbox{\scriptsize{A}}}$ & Auxiliary frame-wise audio prediction head\\
        $(\tilde{x}^t_{ij},\tilde{y}^t_{ij})$ & Coordinates of pixel patch $t$ & $f^{\mbox{\scriptsize{V}}}_C$ & Auxiliary visual prediction head\\
        $o_{ij}$ & Fixed offset for coordinate $(i,j)$ & \multicolumn{2}{l}{\textbf{Hyperparameters}}\\
        $\tilde{e}^{\mbox{\scriptsize{G}}}_t$ & AV-TeST embedded visual token  $(i,j)$ & Symbol & Definition\\
        $\hat{e}^{\mbox{\scriptsize{G}}}_t$ & Reconstructed AV-TeST embedded visual token & $T_G$ & Visual temporal glance limit \\
        $p^{'}_t$ & Softmax prediction of $f^{\mbox{\scriptsize{AV}}}_C$ with feature only at time $t$ & $k$ & Number of selected frames for prediction \\
        $\tilde{s}_t$ & Pseudo-label saliency score & $P$ & Patch size\\
        $p_t$ & Class prediction 
    \end{tabular}}
    \vspace{0.1mm}
    \caption{Table of Notation}
    \label{tab:notation_list}
\end{table*}
\subsection{Patch Extraction Network.}
We explain in detail the process inside the spatial patch extraction network. To enable end-to-end training, we adopt the differentiable solution proposed in \cite{Wang2022_adafocusv2} to obtain $\tilde{\bm{v}}_t$. Suppose that the size of the original frame ${\bm{v}}_t$ and the patch $\tilde{\bm{v}}_t$ is $H\!\times\!W$ and $P\!\times\!P\ (P\!<\!H, W)$, respectively\footnote{In our implementation, the height/width/coordinates are correspondingly normalized using the linear projection $[0,H]\!\to\![0,1]$ and $[0,W]\!\to\![0,1]$. Here we use the original values for the ease of understanding.}. We assume that $\pi$ outputs the continuous centre coordinates $(\tilde{x}^t_{\textnormal{c}}, \tilde{y}^t_{\textnormal{c}})$ of $\tilde{v}_t$ using audio-enhanced global visual feature up to $t^{\textnormal{th}}$ ($\{{e}^{\textnormal{GA}}_{1},\ldots, {e}^{\textnormal{GA}}_{t}\}$), 
\begin{equation}
    \label{eq:cetre_xy}
    \begin{split}
        &(\tilde{x}^t_{\textnormal{c}},\ \tilde{y}^t_{\textnormal{c}}) = \pi(\{{e}^{\textnormal{GA}}_{1},\ldots, {\bm{e}}^{\textnormal{GA}}_{t}\}), \\
        \tilde{x}^t_{\textnormal{c}} \in [&\frac{P}{2}, W-\frac{P}{2}],\ \ \ \tilde{y}^t_{\textnormal{c}} \in [\frac{P}{2}, H-\frac{P}{2}],
    \end{split}
\end{equation}
We refer to the coordinates of the top-left corner of the frame as $(0,0)$, and Eq. (\ref{eq:cetre_xy}) ensures that $\tilde{v}_t$ will never go outside of $v_t$. 

The feed-forward process involves the bilinear interpolation method to enable backpropagation through $(\tilde{x}^t_{\textnormal{c}},\ \tilde{y}^t_{\textnormal{c}})$. As mentioned in the paper, the coordinates of a pixel in the patch $\tilde{v}_t$ can be expressed as the addition of $(\tilde{x}^t_{\textnormal{c}}, \tilde{y}^t_{\textnormal{c}})$ and a fixed offset:
\begin{equation}
    \label{eq:c_plus_offset}
    \begin{split}
        &(\tilde{x}^t_{ij},\ \tilde{y}^t_{ij}) = (\tilde{x}^t_{\textnormal{c}},\ \tilde{y}^t_{\textnormal{c}}) + o_{ij}, \\
        o_{ij} &\in {\left\{ -\frac{P}{2}, -\frac{P}{2} + 1, \ldots, \frac{P}{2} \right\}}^2.
    \end{split}
\end{equation}
$(\tilde{x}^t_{ij}, \tilde{y}^t_{ij})$ denotes the corresponding horizontal and vertical coordinates in the original frame ${v}_t$ to the $i^{\textnormal{th}}$ row and $j^{\textnormal{th}}$ column of $\tilde{v}_t$, while the offset $o_{ij}$ is the vector from the patch center $(\tilde{x}^t_{\textnormal{c}}, \tilde{y}^t_{\textnormal{c}})$ to this pixel. Given a fixed patch size, $o_{ij}$ is a constant conditioned only on $i,j$, regardless of $t$ or the inputs of $\pi$.

Since the values of $(\tilde{x}^t_{\textnormal{c}}, \tilde{y}^t_{\textnormal{c}})$ are continuous, there does not exist a pixel of ${\bm{v}}_t$ exactly located at $(\tilde{x}^t_{ij}, \tilde{y}^t_{ij})$ to directly get the pixel value. Hence, we utilize the four adjacent pixels of $(\tilde{x}^t_{ij}, \tilde{y}^t_{ij})$ to obtain the pixel value using bilinear interpolation. We denote the four surrounding coordinates as $(\lfloor\tilde{x}^t_{ij}\rfloor, \lfloor\tilde{y}^t_{ij}\rfloor)$,  $(\lfloor\tilde{x}^t_{ij}\rfloor\!+\!1, \lfloor\tilde{y}^t_{ij}\rfloor)$,  $(\lfloor\tilde{x}^t_{ij}\rfloor, \lfloor\tilde{y}^t_{ij}\rfloor\!+\!1)$ and $(\lfloor\tilde{x}^t_{ij}\rfloor\!+\!1, \lfloor\tilde{y}^t_{ij}\rfloor\!+\!1)$, respectively, where $\lfloor\cdot\rfloor$ denotes the rounding-down operation. By assuming that the corresponding pixel values of these four pixels are $(m^t_{ij})_{00}$, $(m^t_{ij})_{01}$, $(m^t_{ij})_{10}$, and $(m^t_{ij})_{11}$, the pixel value at $(\tilde{x}^t_{ij}, \tilde{y}^t_{ij})$ (referred to as $\tilde{m}^t_{ij}$) can be obtained via differentiable bilinear interpolation:
\begin{equation}
    \label{eq:bilinear}
    \begin{split}
        \tilde{m}^t_{ij} &= (m^t_{ij})_{00}(\lfloor\tilde{x}^t_{ij}\rfloor\!-\!\tilde{x}^t_{ij}\!+\!1)(\lfloor\tilde{y}^t_{ij}\rfloor\!-\!\tilde{y}^t_{ij}\!+\!1) \\
        &+(m^t_{ij})_{01}(\tilde{x}^t_{ij}\!-\!\lfloor\tilde{x}^t_{ij}\rfloor)(\lfloor\tilde{y}^t_{ij}\rfloor\!-\!\tilde{y}^t_{ij}\!+\!1) \\
        &+(m^t_{ij})_{10}(\lfloor\tilde{x}^t_{ij}\rfloor\!-\!\tilde{x}^t_{ij}\!+\!1)(\tilde{y}^t_{ij}\!-\!\lfloor\tilde{y}^t_{ij}\rfloor) \\
        &+(m^t_{ij})_{11}(\tilde{x}^t_{ij}\!-\!\lfloor\tilde{x}^t_{ij}\rfloor)(\tilde{y}^t_{ij}\!-\!\lfloor\tilde{y}^t_{ij}\rfloor).
    \end{split}
\end{equation}
Consequently, we can obtain the image patch $\tilde{\bm{v}}_t$ by traversing all possible $i,j$ in Eq. (\ref{eq:bilinear}).

Assume we have the training loss $\mathcal{L}$, we can compute the gradient ${\partial\mathcal{L}}/{\partial\tilde{m}^t_{ij}}$ with standard back-propagation. Following the chain rule, we have 
\begin{equation}
        \label{eq:bp_1}
        \frac{\partial\mathcal{L}}{\partial\tilde{x}^t_{\textnormal{c}}} \!=\!\! \sum_{i,j}\! \frac{\partial\mathcal{L}}{\partial\tilde{m}^t_{ij}}
        \frac{\partial\tilde{m}^t_{ij}}{\partial\tilde{x}^t_{\textnormal{c}}}, \ \ \ 
        \frac{\partial\mathcal{L}}{\partial\tilde{y}^t_{\textnormal{c}}} \!=\!\! \sum_{i,j}\! \frac{\partial\mathcal{L}}{\partial\tilde{m}^t_{ij}}
        \frac{\partial\tilde{m}^t_{ij}}{\partial\tilde{y}^t_{\textnormal{c}}}.
\end{equation}
Combining Eq. (\ref{eq:c_plus_offset}) and Eq. (\ref{eq:bp_1}), we can further derive
\begin{equation}
    \label{eq:bp_2}
    \frac{\partial\tilde{m}^t_{ij}}{\partial\tilde{x}^t_{\textnormal{c}}}\!=\!\frac{\partial\tilde{m}^t_{ij}}{\partial\tilde{x}^t_{ij}},\ \ \ 
    \frac{\partial\tilde{m}^t_{ij}}{\partial\tilde{y}^t_{\textnormal{c}}}\!=\!\frac{\partial\tilde{m}^t_{ij}}{\partial\tilde{y}^t_{ij}}.
\end{equation}
Given that $\tilde{x}^t_{\textnormal{c}}$ and $\tilde{y}^t_{\textnormal{c}}$ are the outputs of the network $\pi$, the back-propagation process is able to proceed in an ordinary way.

\begin{figure*}[t]
\centering
    \includegraphics[width=0.8\textwidth]{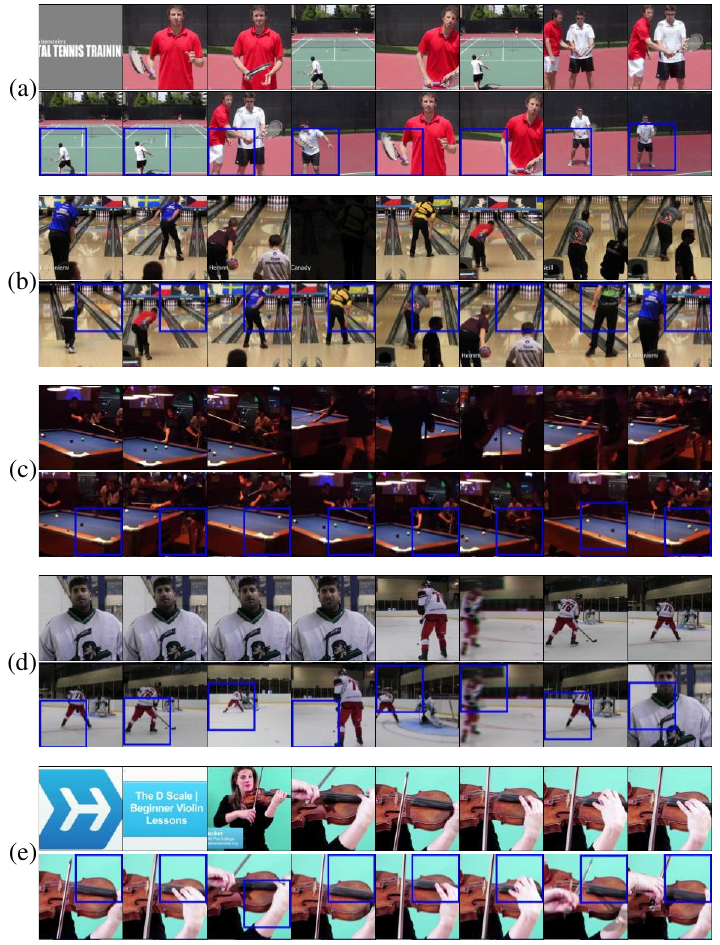}
    \caption{\textbf{Extended qualitative result} shows pair of the first 8 frames in original sequence and the Top-8 salient frames from classes (a) ``\texttt{tennis serve}'', (b) ``\texttt{playing ten pins}'', (c) ``\texttt{playing pool}'',  (d) ``\texttt{playing ice hockey}'', and (e) ``\texttt{playing violin}''. We also provide qualitative results in video format to better comprehend the effect of the audio.}
    \vskip -0.1in
    \label{fig:qual_result_extend}
    \vskip -0.1in
\end{figure*}
\section{Qualitative Results}
We present more qualitative results in image format in Fig. \ref{fig:qual_result_extend} and in video format. Our qualitative results show how the model is able to estimate the salient frames and prioritize them over the non-relevant ones, \textit{e.g.} in (c) salient frames are the ones containing ice hockey-related actions and in (c) and (a) frames with only text are non-salient. From the examples in video format, we observe how strong audio cues are present in the salient frames. For example, in ``\texttt{playing ten pins}'' class sample, the sound of the ball crashing the pins provide strong cues to estimate saliency.